\def\year{2021}\relax
\documentclass[letterpaper]{article} 
\usepackage{aaai21}  
\usepackage{times}  
\usepackage{helvet} 
\usepackage{courier}  
\usepackage[hyphens]{url}  
\usepackage{graphicx} 
\usepackage{natbib}  
\usepackage{caption} 
\usepackage{algorithm}
\usepackage[noend]{algpseudocode}
\usepackage{times}
\usepackage{latexsym}
\usepackage{diagbox}
\usepackage{tikz}
\usepackage{microtype}
\usepackage{graphicx}
\usepackage{booktabs} 
\usepackage{subcaption}
\usepackage{outlines}
\usepackage{hyperref}
\usepackage{multirow}
\usepackage{amsmath}
\usepackage{amsfonts} 
\usepackage{hyperref}

\usetikzlibrary{fit,shapes.geometric, positioning, arrows, decorations.pathreplacing, calc, arrows.meta}
\def\checkmark{\tikz\fill[scale=0.4](0,.35) -- (.25,0) -- (1,.7) -- (.25,.15) -- cycle;}

\usepackage{microtype}

\DeclareMathOperator*{\argmax}{arg\,max}

\title{One-shot Learning for Temporal Knowledge Graphs}
\author {
    Mehrnoosh Mirtaheri,\textsuperscript{\rm 1}
    Mohammad Rostami,\textsuperscript{\rm 1}
    Xiang Ren,\textsuperscript{\rm 1}
    Fred Morstatter,\textsuperscript{\rm 1}
    Aram Galstyan \textsuperscript{\rm 1} \\
}
\affiliations {
    \textsuperscript{\rm 1} USC Information Sciences Institute \\
    \{mehrnoom, mrostami, xiangren, fredmors, galstyan\}@isi.edu
}

\begin{document}

\maketitle

\begin{abstract}
Most real-world knowledge graphs are characterized by a long-tail relation frequency distribution where a significant fraction  of relations occurs only a handful of times. This observation has given rise to recent interest in low-shot learning methods that are able to generalize from only a few examples. The existing approaches, however, are tailored to static knowledge graphs and not easily generalized to temporal settings, where data scarcity poses even bigger problems, e.g., due to occurrence of new, previously unseen relations. We address this shortcoming by proposing a one-shot learning framework for link prediction in temporal knowledge graphs. Our proposed method employs a self-attention mechanism to effectively encode temporal interactions between entities, and  a network  to compute a similarity score between a given query and a (one-shot) example. Our experiments show that the proposed algorithm outperforms the state of the art baselines for two well-studied benchmarks while achieving significantly better performance for sparse relations.

\end{abstract}

\section{Introduction}
Large scale knowledge graphs (KGs) have become a crucial component for performing various Natural Language Processing (NLP) tasks, including cross-lingual translation \cite{wang2018cross}, Q\&A~\cite{yao2014information} and relational learning~\cite{nickel2016review}. Despite being effective, KGs typically suffer from incompleteness; therefore automatic KG completion is crucial    for the above reasoning tasks. 

 Previous methods of KG completion have traditionally focused on learning representations over static knowledge graphs such as YAGO~\cite{kasneci2009yago} and WikiData~\cite{vrandevcic2014wikidata}. 
 Due to the rapid growth of event datasets, automatically extracted from news archives, there has also been a significant recent interest in learning for Temporal Knowledge Graphs (TKG). Recent attempts on learning over TKGs mostly focus on predicting either missing events (links between entities) for an observed timestamp ~\cite{dasgupta2018hyte, garcia2018learning, leblay2018deriving}, or future timestamps by leveraging the temporal dependencies between entities~\cite{trivedi2017know, jin2019recurrent}.

\begin{figure}
    \centering
    \includegraphics[width=0.5\textwidth, trim={0 0 0cm 0.8cm}, clip]{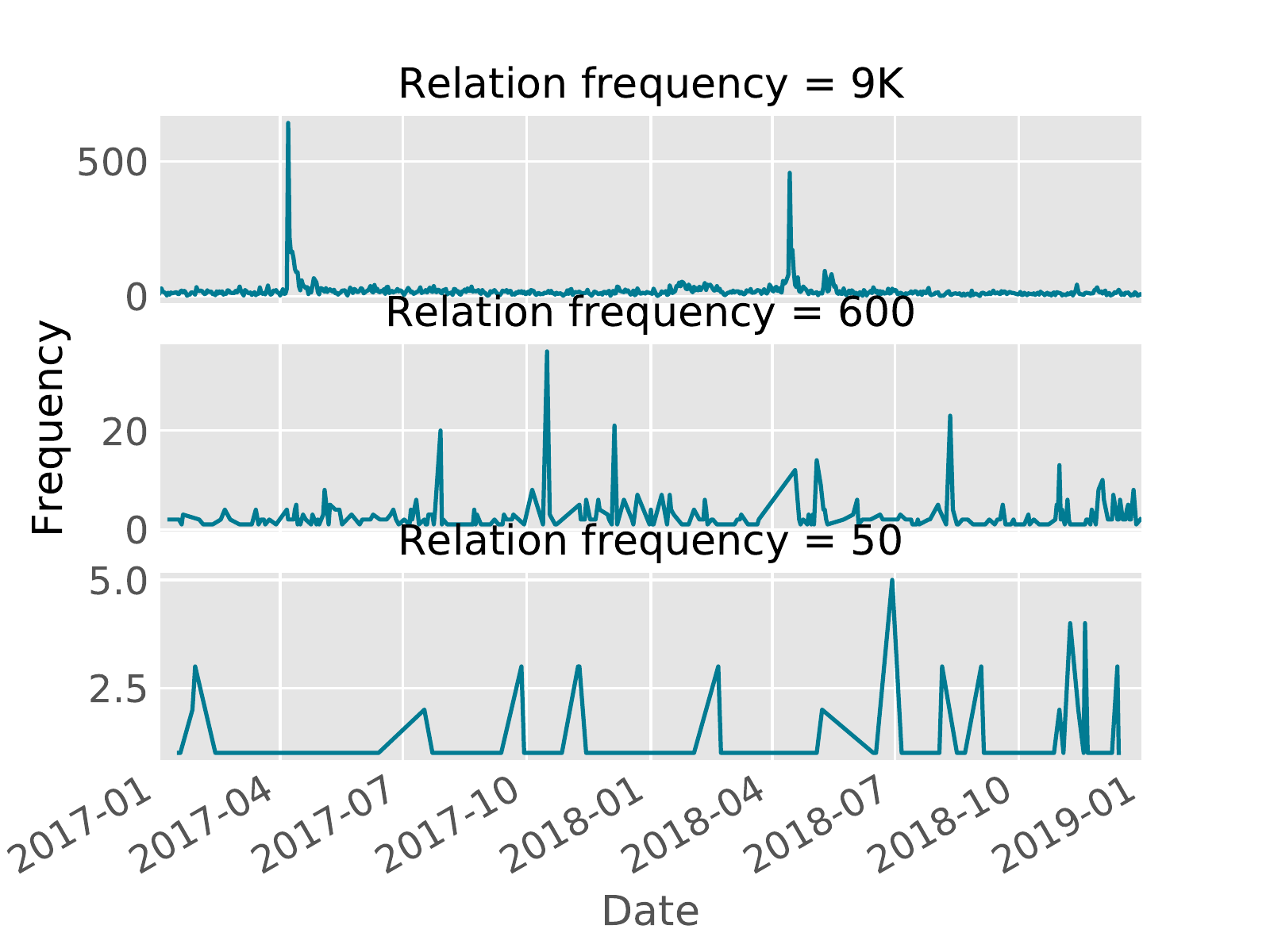}
    \vspace{-0.6cm}
    \caption{Distribution of three relations with different frequency over ICEWS Jan 2017-Jan 2019. The distribution of occurrences is highly heterogeneous.}
    \label{fig:rel_count}
    \vspace{-0.5cm}
\end{figure} 

Most of the existing KG completion methods rely on a sufficiently large number of training examples per relation. Unfortunately, most real-world KGs have a long-tail structure, so many relationships occur only a handful of times. Recent research has addressed the problem by developing efficient low-shot learning methods.  Xiong et al. \citeyear{xiong2018one} pioneered employing few-shot learning (FSL) to infer a link between two entities given only one training example, which is done by learning a matching metric over the embeddings extracted from the one-hop neighborhood of entities in the graph. Several followup works improving upon GMatching \cite{bose2019meta, chen2019meta, wang2019meta, wang2019tackling} are also comprised of an encoder component extracting features for the meta learning component.

The data scarcity issue is exacerbated for temporal graphs, since the dynamic governing the evolution of those graphs might be highly non-stationary. First, new types of relationships/events might emerge that have not been observed before. Furthermore,  even if a given relation has been observed frequently over some time interval, the distribution of occurrences over that interval might be highly inhomogeneous and bursty, as shown in  Fig.~\ref{fig:rel_count}. 
The existing methods, which are developed for static graphs, cannot account for such dynamics. First, their encoders are not able to incorporate the existing temporal dependencies between entities into the model. Second, their task definition in the FSL framework does not account for time constraints. Thus, we need novel low-shot learning approaches to accurately model TKGs.

{We address this challenge by proposing a new model, Few-shot Temporal Attention Graph Learning (FTAG), to effectively infer the true entity pairs for a relation, given only one support entity pair. Our model learns a representations for entities by aggregating temporal information over neighborhoods. It employs a self-attention mechanism to sequentially extract an entity's neighborhood information over time. Temporally adjacent events could convey useful information about the events that will happen in the future. Thus, self-attention, as a powerful tool for sequence modeling, has been used to extract a time-aware representation from the entity neighborhood.
Then, the learned representation would be used to match a given support entity pair with a query entity pair and generate a similarity score proportional to the likelihood of the event.}

Our contributions are summarized as follows: 

 \begin{itemize}
     \item We introduce a one-shot learning framework for temporal knowledge graphs, which generalizes over existing low-shot techniques for static graphs.   
    
     \item We propose a temporal neighborhood encoder with a  self-attention mechanism that effectively extracts the temporally-resolved neighborhood information for each entity in the graph. 
     
    \item We conduct extensive experiments with two popular real-world datasets and demonstrate the superiority of the proposed model over state of the art baselines.  
    
    \item We construct two new publicly-available benchmarks for one-shot learning over temporal knowledge graphs.
 \end{itemize}{}

\section{Problem Formulation}

We present the formal definition of a TKG and explore  few-shot learning tasks for temporal link prediction. 

\subsection{Temporal Knowledge Graph Completion}
A TKG can be represented as a set of quadruples $G = \{(s, r, o, t) | s, o \in \mathcal{E}, r \in \mathcal{R} \}$, where $\mathcal{E}$ is the set of entities, $\mathcal{R}$ is the set of relations and $t$ is the timestamp. Graph completion for a static KG involves predicting new facts by either predicting an unseen object entity for a given subject and relation $(s, r, ?)$ or predicting an unseen link between the subject and object entity $(s, ? , o)$. In this work, we are interested in the former case, but at a particular timestamp $t$. More formally, we want to predict new events $(s, r, ?, t)$ at time $t$ by ranking the true object entities higher than others by a scoring function $\mathcal{P}$, parameterized by a neural network and its value indicating the event likelihood. 
The key idea in  modeling the temporal events is that an event likelihood depends on the events in the previous $\ell$ timesteps $\{t-\ell, \dots, t-2, t-1\}$. In Section \ref{sec:model_encoder} we explain in detail that how temporal information is encoded in our model. 

\subsection{Few-shot Learning and Episodic Training}
Few-shot Learning (FSL) focuses on building and training a model with only a few labeled instances for each class. Meta-learning is a framework to address FSL where we incorporate a large set of tasks, and each task mimicks an \textit{N-way K-shot} classification scenario. The aim is to leverage the shared information across the tasks to compensate for the scarcity of information about each task resulting from having few labeled data points. 

The idea of episodic training for meta-learning is to match the training procedure with the inference at test time~\cite{vinyals2016matching}. More specifically, consider that we have a large set of tasks $\mathcal{T}$. Each episode consists of a subset of tasks $L$, sampled from $\mathcal{T}$, a support set $S$, and a batch $B$ both sampled from $L$. Note that both $S$ and $B$ are labeled examples, labels coming from $L$. The goal is to train a model that maps the few examples in the support set into a classifier. The probabilistic optimization objective for this problem can be formulated as:

\small
\begin{equation}
\theta = \underset{\theta}{\argmax} E_{L \sim \mathcal{T}} \Bigg[ E_{S\sim L, B \sim L}\Bigg[ \underset{(x, y) \in B}{\sum} \log P_{\theta}(y | x, S) \Bigg]\Bigg].
\label{temp}
\end{equation}
 \normalsize
 
We adopt the standard episodic training framework in Eq.~\eqref{temp} for the purpose of TKG completion.

\subsection{Few-shot TKG Setup}
\label{sec:fewshot-setup}
As mentioned earlier, data scarcity is even a bigger problem in relational learning with TKGs. Few-shot episodic training has been proven to be effective to tackle this problem for static KGs~\cite{xiong2018one}. We further extend the framework proposed by Xiong et al.~\citeyear{xiong2018one} for TKG completion. 

Given a TKG, $G = \{(s, r, o, t) | s, o \in \mathcal{E}, r \in \mathcal{R} \}$, the relations of $\mathcal{R}$ are divided into two groups based on their frequency: frequent relations $\mathcal{F}$ and sparse relations $\mathcal{T}$. The sparse relations are used to build the task set needed by the model for the episodic training. Each task is defined as a sparse relation $r \in \mathcal{T}$ and has its own training and test set, denoted as support and query set and defined as: 
\begin{equation}
\begin{split}
&S_r = \{(s_0, r, o_0, t_0) | s_0, o_0\in\mathcal{E}\}\\
&Q_r = \{(s_q, r, o_q, t_q) | s_q, o_q\in\mathcal{E}\},
\end{split}
\end{equation}
 
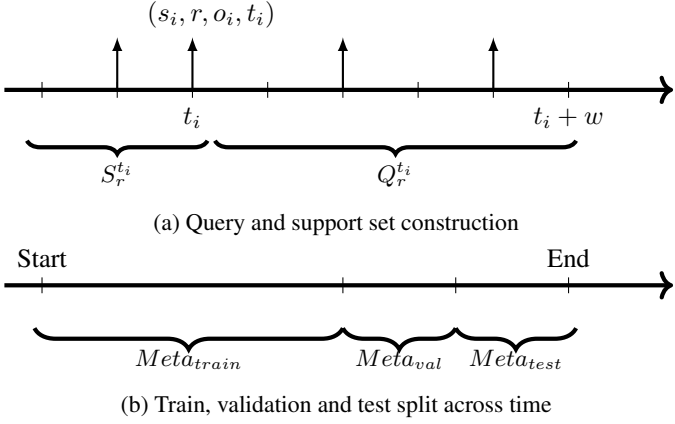
\begin{figure}
    \centering
    
    

    


    
    \begin{subfigure}[t]{0.5\textwidth}
        \begin{tikzpicture}[scale=1, transform shape]
    \draw[ultra thick, ->] (0,0) -- (\textwidth,0);
    
    \foreach \x in {0.5,1.5,2.5,3.5,4.5,5.5,6.5,7.5}
    \draw (\x cm,3pt) -- (\x cm,-3pt);
    
    \draw[thick, -latex] (1.5, 0) -- (1.5, 0.7);
    \draw[thick, -latex] (2.5, 0) -- (2.5, 0.7)node[xshift=0.25cm, above,thick] {$(s_i, r, o_i, t_i)$ };
    \draw[thick, -latex] (4.5, 0) -- (4.5, 0.7);
    \draw[thick, -latex] (6.5, 0) -- (6.5, 0.7);
    
    \draw[ultra thick] (7.5,0) node[below=3pt,thick] {$t_i + w$ } node[above=3pt] {};
    \draw[ultra thick] (2.5,0) node[below=3pt,thick] {$t_i$} node[above=3pt] {};

    \draw [ black, ultra thick,decorate,decoration={brace,mirror, amplitude=5pt},
           yshift=-5pt] (2.8,-0.5) -- (7.6,-0.5)
           node [black,midway,below=4pt] {\footnotesize  $Q_r^{t_i}$};
           
    \draw [ black, ultra thick,decorate,decoration={brace,mirror, amplitude=5pt},
           yshift=-5pt] (0.3,-0.5) -- (2.7,-0.5)
           node [black,midway,below=4pt] {\footnotesize $S_r^{t_i}$};
    
    \end{tikzpicture}
    \caption{Query and support set construction}
    \label{fig:query}
    \end{subfigure}
    
    \begin{subfigure}[t]{0.5\textwidth}
        \begin{tikzpicture}[scale=1, transform shape]
    \draw[ultra thick, ->] (0,0) -- (\textwidth,0);
    
    \foreach \x in {0.5,4.5,6,7.5}
    \draw (\x cm,3pt) -- (\x cm,-3pt);
    
    \draw[ultra thick] (0.5,0) node[above=3pt,thick] {Start} node[above=3pt] {};
    \draw[ultra thick] (7.5,0) node[above=3pt,thick] {End} node[above=3pt] {};

    \draw [ black, ultra thick,decorate,decoration={brace,mirror, amplitude=8pt},
           yshift=-2pt] (0.4,-0.5) -- (4.5,-0.5)
           node [black,midway,below=4pt] {\footnotesize  $Meta_{train}$};
           
    \draw [ black, ultra thick,decorate,decoration={brace,mirror, amplitude=8pt},
          yshift=-2pt] (4.5,-0.5) -- (6,-0.5)
          node [black,midway,below=4pt] {\footnotesize $Meta_{val}$};
    
    \draw [ black, ultra thick,decorate,decoration={brace,mirror, amplitude=8pt},
           yshift=-2pt] (6,-0.5) -- (7.6,-0.5)
           node [black,midway,below=4pt] {\footnotesize $Meta_{test}$};
    \end{tikzpicture}
    \caption{Train, validation and test split across time}
    \label{fig:split}
    \end{subfigure}
    \caption{(a) The query set is selected from $[t_i, t_i + w]$, with $t_i$ being the last quads timestamp in the support (b) There is no time overlap between quadruples in validation and test.}
    \label{fig:my_label}
    \vspace{-1em}
\end{figure}

Where $S_r$ contains one labeled example. At each episode, one relation is selected at random and one quadruple containing that relation to form the support set. We can select the quadruples for the query set in two ways: 
\begin{enumerate}
    \item \textbf{Random}: At each episode, $m$ quadruples are selected randomly for the query set.  
    \item \textbf{Time dependent}: The quadruples of the query set are restricted by their distance from the support set timestamp: 
    \vspace{-0.4cm}
    \begin{equation}
        Q_r^t = \{(s_q, r, o_q, t_q) | s_q, o_q\in\mathcal{E}, t_q \in [t_0, t_0 + w]\}
    \end{equation}
    Where $t_0$ is the support set timestamp. Figure~\ref{fig:query} illustrates  the time constraint used for selecting the query examples. More details on sampling procedure for the support and query set are provided in Section~\ref{sec:training}. The parameter $w$ is called the \textit{episode length}.
\end{enumerate}


The loss function $l_{\theta}$ at each episode optimizes a score function $\mathcal{P_{\theta}}$ such that for a given test query in $Q_r^t$, the true object entities are ranked higher than the others. The score function is a metric space learnt during the training, and the score represents the  similarity between a test query and the support set representation. The  final optimization loss   is:
\begin{equation}
\mathcal{L} = E_{r \sim \mathcal{T} } \bigg[ E_{Q_r^t \sim G, S_r^t \sim G} \bigg[ l_{\theta}(Q_r^t | S_r^t)\bigg] \bigg] 
\end{equation}

{The relations in $\mathcal{T}$ are divided into mutually exclusive sets: $\mathcal{T}_{meta-train}, \mathcal{T}_{meta-test}, \mathcal{T}_{meta-val}$. From this, $meta_{train}$ is defined as:
\[meta_{train} = \{(s, r, o, t) | r \in \mathcal{T}_{meta-train}\}\]
$meta_{val}$ and $meta_{test}$ are defined similarly. To make the temporal setup more representative of the real world, we do not allow any time overlap between the quadruples in $meta_{train}$, $meta_{val}$ and $meta_{test}$. Figure \ref{fig:split} depicts the time split for $meta_{train}$, $meta_{val}$ and $meta_{test}$. To have the episodes in the training match with the inference, the split window for validation and test is equal to $w$.}


Finally, we assume that the model has access to a background knowledge graph defined as $G' = \{(s, r, o, t) | s, o\in\mathcal{E}, r \in \mathcal{F}\}$, and the entity set $\mathcal{E}$ is a closed set, i.e., there are no unseen entities during the inference time. 

\section{Model}
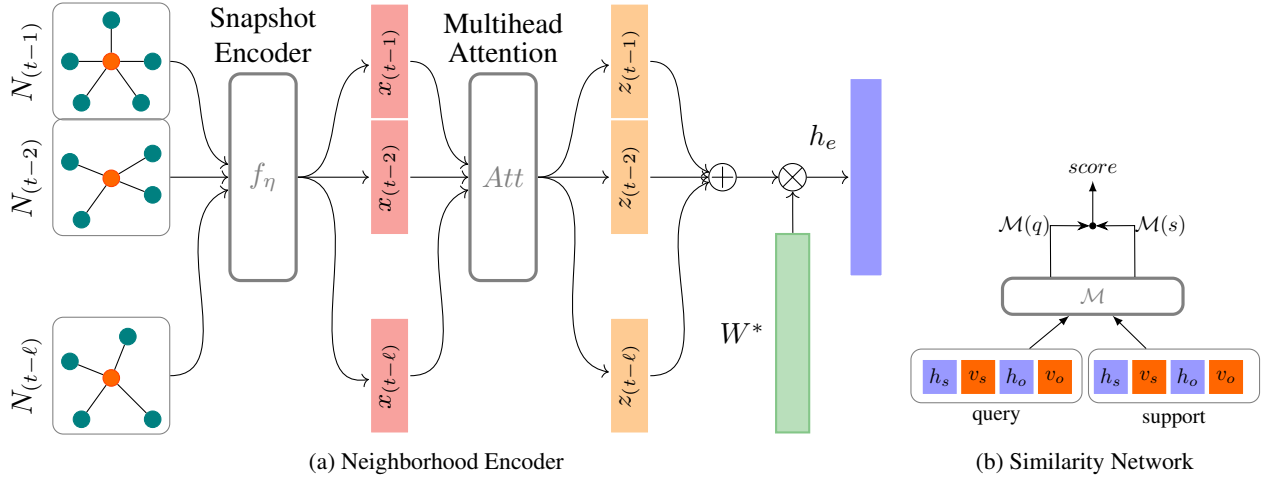
\begin{figure*}[t!]
    \centering
    \begin{subfigure}[t]{0.65\textwidth}
        \centering
        \begin{tikzpicture}[scale=1.1, transform shape]
        \definecolor{paramcolor}{HTML}{50ae55}
        \definecolor{inputcolor}{HTML}{fd9727}
        \definecolor{outputcolor}{HTML}{f1453d}
        \tikzset{
         main node/.style={anchor=base,circle,fill=teal!100,draw=teal!100,minimum size=0.2cm,inner sep=0.5pt}, 
        middle node/.style={anchor=base,circle,fill=orange!120,draw=orange!150,minimum size=0.2cm,inner sep=0.5pt},
    	attention/.style = {rectangle, draw, rounded corners, very thick, gray, minimum width=0.8cm, minimum height = 2.5cm},
    	outputvec/.style = {rectangle, draw=outputcolor!0, fill=outputcolor!50, thick, minimum width=0.1cm, minimum height = 1.4cm},
    	attoutput/.style = {rectangle, draw=outputcolor!0, fill=inputcolor!50, thick, minimum width=0.1cm, minimum height = 1.4cm},
    	parameter/.style = {rectangle, draw=paramcolor!70, fill=paramcolor!40, thick, minimum width=0.4cm, minimum height = 2.4cm},
    	history/.style = {rectangle, draw=blue!0, fill=blue!40, thick, minimum width=0.4cm, minimum height = 2.4cm},
    	do path picture/.style={%
        path picture={%
          \pgfpointdiff{\pgfpointanchor{path picture bounding box}{south west}}%
            {\pgfpointanchor{path picture bounding box}{north east}}%
          \pgfgetlastxy\x\y%
          \tikzset{x=\x/2,y=\y/2}%
          #1
        }
      },
      sin wave/.style={do path picture={    
        \draw [line cap=round] (-3/4,0)
          sin (-3/8,1/2) cos (0,0) sin (3/8,-1/2) cos (3/4,0);
      }},
      cross/.style={do path picture={    
        \draw [line cap=round] (-1,-1) -- (1,1) (-1,1) -- (1,-1);
      }},
      plus/.style={do path picture={    
        \draw [line cap=round] (-3/4,0) -- (3/4,0) (0,-3/4) -- (0,3/4);
      }}
}

        \node[main node] (1) at (-0.5,0.5){};
        \node[middle node] (2) at (0,0.5){};
        \node[main node] (3) at (0.5,0.5) {};
        \node[main node] (4) at (-0.360,0){};
        \node[main node] (5) at (0.360,0){} ;
        \node[main node] (6) at (0,1){};
    
        \draw (1) -- (2);
        \draw (4) -- (2);
        \draw (6) -- (2);
        \draw (2) -- (3);
        \draw (2) -- (5);
        
        \node[middle node, below=1.2 of 2] (7){};
        \node[main node, below=1.4 of 3] (8) {};
        \node[main node, below=1 of 1] (9){};
        \node[main node, below=1.4 of 6, xshift=0.5cm] (10){} ;
        \node[main node, below=1.2 of 4] (11){};
    
        \draw (8) -- (7);
        \draw (9) -- (7);
        \draw (10) -- (7);
        \draw (7) -- (11);
        
        \node[middle node,below=2.2 of 7] (12)  {};
        \node[main node,below=2.5 of 8] (13) {};
        \node[main node,below=2.2 of 9] (14) {};
        \node[main node,below=2 of 10, xshift=-0.3cm] (15) {} ;
        \node[main node,below=2.2 of 11] (16) {};
        
        \draw (13) -- (12);
        \draw (14) -- (12);
        \draw (15) -- (12);
        \draw (16) -- (12);
        
        \node[rectangle, draw, rounded corners, thin, gray, label={left:\rotatebox{90}{$N_{(t-1)}$}},
        fit={(-.6, -0.1) (-0.6,1.1) (0.6,-0.1) (0.6,1.1)}] (G1){};
        \node[rectangle, draw, rounded corners, thin, gray,label={left:\rotatebox{90}{$N_{(t-2)}$}},
        fit={(-.6, -1.5) (-0.6,-0.3) (0.6,-1.5) (0.6,-0.3)}](G2){};
        \node[rectangle, draw, rounded corners, thin, gray,label={left:\rotatebox{90}{$N_{(t-\ell)}$}},
        fit={(-.6, -3.9) (-0.6,-2.7) (0.6,-3.9) (0.6,-2.7)}] (G3){};
        
        \node[attention, right=0.7 cm of G2, label={above:\shortstack{Snapshot \\ Encoder}}](snap){$f_{\eta}$};
        
        \draw[->](G1) to[out=0,in=160,looseness=1] (snap);
        \draw[-{>}](G2) to[out=0,in=180,looseness=1] (snap);
        \draw[-{>}](G3) to[out=0,in=-160,looseness=1](snap);
        
        \node[outputvec, right=2.4cm of G1](o1){\small \rotatebox{90}{$x_{(t-1)}$}};
        \node[outputvec, right=2.4cm of G2](o2){\small \rotatebox{90}{$x_{(t-2)}$}};
        \node[outputvec, right=2.4cm of G3](o3){\small \rotatebox{90}{$x_{(t-\ell)}$}};
        
        \draw[-{>}](snap) to[out=0,in=190,looseness=1] (o1);
        \draw[-{>}](snap) to[out=0,in=180,looseness=1] (o2);
        \draw[-{>}](snap) to[out=0,in=190,looseness=1] (o3);
        
        \node[attention, right=0.7 of o2,label={above:\shortstack{Multihead \\ Attention}}](att){$Att$};

        \draw[-{>}](o1) to[out=0,in=160,looseness=1] (att);
        \draw[-{>}](o2) to[out=0,in=180,looseness=1] (att);
        \draw[-{>}](o3) to[out=0,in=-160,looseness=1] (att);
        
        \node[attoutput, right=5.3cm of G1](z1){\small \rotatebox{90}{$z_{(t-1)}$}};
        \node[attoutput, right=5.3cm of G2](z2){\small \rotatebox{90}{$z_{(t-2)}$}};
        \node[attoutput, right=5.3cm of G3](z3){\small \rotatebox{90}{$z_{(t-\ell)}$}};
        
        \draw[-{>}](att) to[out=0,in=190,looseness=1] (z1);
        \draw[-{>}](att) to[out=0,in=180,looseness=1](z2);
        \draw[-{>}](att) to[out=0,in=-190,looseness=1] (z3);
        
        \node [circle, draw, plus, right=0.7cm of z2](concat1){};
        \draw[-{>}](z1) to[out=0,in=160,looseness=1] (concat1);
        \draw[-{>}](z2) to[out=0,in=180,looseness=1] (concat1);
        \draw[-{>}](z3) to[out=0,in=-160,looseness=1] (concat1);
        
        \node [circle, draw, cross, right=0.5 of concat1](cross1){};
        \draw[-{>}](concat1) to[out=0,in=180,looseness=1] (cross1);
        
        \node [parameter, below= 0.5 of cross1, label={left:$W^*$}](param){};
        \draw[-{>}](param) to[out=90,in=-90,looseness=1] (cross1);
        
        \node [history, right= 0.5 of cross1, label={above left:$h_e$}](hist){};
        \draw[-{>}](cross1) to[out=0,in=180,looseness=1] (hist);

    \end{tikzpicture}
        \caption{Neighborhood Encoder}
        \label{fig:ex_graph_a}
    \end{subfigure}%
    ~
   \begin{subfigure}[t]{0.3\textwidth}
   \centering
    \begin{tikzpicture}[scale=0.8, transform shape]
    \definecolor{paramcolor}{HTML}{50ae55}
    \definecolor{inputcolor}{HTML}{fd9727}
    \definecolor{outputcolor}{HTML}{f1453d}
        \tikzset{
        histvec/.style = {rectangle, draw=blue!0,fill=blue!40, thick, minimum width=0.6cm, minimum height = 0.6cm},
    	embvec/.style = {rectangle, draw=orange!0, fill=orange!120, thick, minimum width=0.6cm, minimum height = 0.6cm},
    	module/.style = {rectangle, draw, rounded corners, very thick, gray, minimum width=3cm, minimum height = 0.6cm},
        do path picture/.style={%
        path picture={%
          \pgfpointdiff{\pgfpointanchor{path picture bounding box}{south west}}%
            {\pgfpointanchor{path picture bounding box}{north east}}%
          \pgfgetlastxy\x\y%
          \tikzset{x=\x/2,y=\y/2}%
          #1
        }
      },
      plus/.style={do path picture={    
        \draw [line cap=round] (-3/4,0) -- (3/4,0) (0,-3/4) -- (0,3/4);
      }},
      cross/.style={do path picture={    
        \draw [line cap=round] (-1,-1) -- (1,1) (-1,1) -- (1,-1);
      }},
      bullet/.style={circle, fill, minimum size=2pt,
              inner sep=1pt, outer sep=0pt}
      }
        
        \node[histvec, below left=0.5cm](hsq){$h_s$};
        \node[embvec, right=0.01cm of hsq](vsq){$v_s$};
        \node[histvec, right=0.01cm of vsq](hoq){$h_o$};
        \node[embvec, right=0.01cm of hoq](voq){$v_o$};
        
        \node[histvec, right=0.3cm of voq](hss){$h_s$};
        \node[embvec, right=0.01cm of hss](vss){$v_s$};
        \node[histvec, right=0.01cm of vss](hos){$h_o$};
        \node[embvec, right=0.01cm of hos](vos){$v_o$};
        
        \node[rectangle, draw, rounded corners, thin, gray, label={below:query},
        fit={(-1, -.95) (-1,-.35) (1.55,-.95) (1.55,-.35)}] (query){};
        
        \node[rectangle, draw, rounded corners, thin, gray, label={below:support},
        fit={(1.95, -.95) (1.95,-.35) (4.5,-.95) (4.5,-.35)}] (support){};
        
        \node[module, above right =0.5cm](lstm){$\mathcal{M}$};
        \node[bullet, draw, above=0.8 of lstm](concat1){};
        \node[above=0.7 of concat1](score){$score$};
        
        \draw[-latex](query) -- (lstm);
        \draw[-latex](support) -- (lstm);

        \path [draw=black, -Stealth] 
            ([xshift=-.7cm]lstm.north) -- (1.17, 1.85) -- (concat1) node[midway,left, xshift=-.2cm] {$\mathcal{M}(q)$};
        \path [draw=black, -Stealth] 
            ([xshift=.7cm]lstm.north) -- (2.57, 1.85) -- (concat1)
            node[midway,right,xshift=.2cm] {$\mathcal{M}(s)$};
            \draw[-latex](concat1) -- (score);

    \end{tikzpicture}
        \caption{Similarity Network}
        \label{fig:components}
   \end{subfigure}%
    

    \caption{Depiction of the main components of our model: (a) Temporal neighborhood encoder comprised of a snapshot encoder $f_\eta$ and the $Att$ modules; (b) Similarity score is computed via the inner product.} 
    \label{fig:model}
    \vspace{-0.2cm}
\end{figure*}
Our model is built upon two main fundamentals: (i) a representation for the support set and query instances to preserve the relational/sequential graph structure, and (ii) a metric to determine the similarity of the support set and a query instance. From this, our model consists of two main components (Figure \ref{fig:model}), as follows:

\noindent\textbf{Neighborhood Encoder}. The neighborhood encoder represents the neighborhood information of a given entity $e$ as a $d$ dimensional vector $h_e$. It encodes the one-hop neighborhood structure during the past $\ell$ timesteps as a sequence. In Section \ref{sec:model_encoder} we explain the detail of obtaining a test query and support set representation via the encoder. 

\noindent\textbf{Similarity Network}. A similarity function parameterized by a neural network, $\mathcal{M }(q,S))$, that outputs a scalar similarity score between the query instance $q$, and the support set $S$, where $q =(s_q, r, o_q, t)$ is a potential event and the similarity score is proportional to the likelihood of that event.


  
%

\subsection{Neighborhood Encoder} \label{sec:model_encoder} For a given entity $e$, we define $\mathcal{N}^r_\tau(e)$ as the set of all adjacent entities connected to $e$ with relation $r$ at time $\tau$, and the temporal neighborhood $N(e) = \bigcup_{r \in \mathcal{F}, \tau \in [t-\ell, t-1]} N^r_\tau (e)$. The neighborhood encoder is comprised of two parts: (i) function $f_\eta$ that encodes the one-hop neighborhood at a given timestamp $\tau$, and (ii) function $g$, that utilizes the output of function $f_\eta$, at previous timesteps, to generate a temporal neighborhood representation.

\subsubsection{Snapshot Aggregation.} The snapshot aggregator $f_{\eta}$ aggregates local neighborhood information at a specific time $\tau$.
\begin{equation}
\begin{split}
   &f_{\eta}(N_{\tau}(e)) =\\&
\sigma (\frac{1}{C_{e_{\tau}}}\sum_{r\in \mathcal{F}}\sum_{e_j \in N^r_\tau(e)} (W^T[v_r:v_{e_j}] + b))\\&
x_\tau^e = [f_{\eta}(N_{\tau}(e)): v_e],
\end{split}
\end{equation}
where $C_{e_{\tau}}$ is a normalizing factor,  $v_{e_j} \in \mathbf{R}^{d\times1}$, $v_r \in \mathbf{R}^{d\times1}$are entity and relation representations, and $W \in \mathcal{R}^{2d \times d}$ and $b \in \mathcal{R}^{d\times1}$ are model parameters to be learnt. $\sigma(.)$ is a nonlinear activation function ($Relu$ in our case). 

\subsubsection{Sequential Aggregation}
\label{sec:attention}
{Function $g$ aggregates the sequence of snapshots from previous $l$ timesteps $\{t-\ell, \dots, t-2, t-1\}$. We then use a transformer, as proposed in \cite{vaswani2017attention}. This is an encoder-decoder model solely based on attention mechanism that proves to be powerful for modeling sequential data. Here, the encoder part, denoted as \textit{Att},  is employed to effectively capture the time dependencies between the event sequences. The main component of \textit{Att} function is a layer, made up of two sublayers:}

\noindent\textbf{Attention sublayer} projects the input sequence to a query and a set of key-value vectors. 
\begin{equation}
\begin{split}
& \textrm{MultiHead}(Q, K, V ) = [head_1: ...:head_h]W^O \\&
head_i = \textrm{Attention}(QW_i^Q , K W_i^K , V W_i^V ),
\end{split}
\label{eq:attention}
\end{equation}
where $W_i^Q \in \mathbf{R}^{d_{model} \times d_k}$ , $W_i^K \in\mathbf{R}^{d_{model} \times d_k}$ , $W_i^V \in \mathbf{R}^{d_{model} \times d_v}$, $W^O \in \mathbf{R}^{hd_v \times d_{model}}$ are parameter matrices, and $d_{model}$ is the input embedding dimension ($d_{model}=2d$ in our case).

\noindent\textbf{Position wise sublayer} is a fully connected feed-forward network, applied to each sequence position separately and identically.
\begin{equation}
FFN(x_\tau) = \max(0, x_\tau W_1 + b_1 )W_2 + b_2.
\end{equation}

The $\textit{Att}(x, n_{head}, n_{layer})$ takes as input a sequence of neighborhood snapshot representations $x = \{x^e_{t-\ell}, \dots, x^e_{t-1}\}$, the number of layers, and number of attention heads, and maps input sequence $x$ to a time-aware sequence output $z = [z_{t-\ell}, \dots, z_{t-1}]$ as follows:
\begin{equation}
z = \textit{Att}(x, n_{head}, n_{layer}))
\end{equation}

Finally, the temporal neighborhood representation for $e$ at time $t$ is obtained by:
\begin{equation}
    h_e  = \sigma ([z_{t-\ell}: \dots : z_{t-1}] W^*),
\end{equation}
 where $W^* \in \mathbf{R}^{2dl \times d_{out}}$ is a  parameter matrix, $[:]$ is concatenation and $\sigma(.)$ is a nonlinear activation function ($Relu$ in our case). More detail on \textit{Att} is provided in Appendix \ref{sec:app-att}. 

\subsection{Similarity Network} 
\label{sec:metric-learning}
Given the Neighborhood encoder, every pair of subject and object $(s, o)$ can be represented as a vector $[h_s:v_s:h_o:v_o]$, where $h_s$ and $h_o$ are the temporal representations obtained from the neighborhood encoder and $v_s$ and $v_o$ are the embeddings for the subject and the object entities. 



{Given the support entity pair $(s_0, o_0)$ for a relation $r$, we learn the representation for similarity from the support and the query entity pair by two layers of fully connected layers: 
\begin{equation}
\begin{split}
  &  x^{(1)} = \sigma (W^{(1)}x + b^{(1)}) \\&
    x^{(2)} = W^{(2)}x^{(1)} + b^{(2)} \\&
   \mathcal{M}(x) = x^{(2)}+ x.
   \end{split}
\end{equation}
The inner product is used to compute the similarity score between the support and query entity pair:
\begin{equation}
 score = \mathcal{M}(s)^T.\mathcal{M}(q),   
\end{equation}
}
{where $s = [h_{s_0}: v_{s_0}: h_{o_0}: v_{o_0}]$ and $q = [h_{s_q}: v_{s_q}: h_{o_q}: v_{o_q}]$. We use the dot product to output a similarity score between the support  and query pair that corresponds to the likelihood of $s_q$ and $o_q$ being connected with $r$.  }

\begin{table*}[t]
\centering

    \begin{tabular}{l || c c c c | c c c c}
        \toprule
        & \multicolumn{4}{c|}{GDELT} & \multicolumn{4}{c}{ICEWS} \\
        \cline{2-9}
        Model & H@1 & H@5 & H@10 & MRR & H@1 & H@5 & H@10 & MRR\\
        \midrule
        TTransE  &0.025 &0.075 &0.138 &0.060&0.004 & 0.047 & 0.107 & 0.038\\
        TATransE & 0.062 & 0.200 &0.362 &0.151 & 0.084 & 0.238 & 0.418 & 0.168\\
        ReNet  & 0.064 & 0.191&0.319 & 0.146 & 0.126 & 0.289 & 0.407 & 0.209\\
        \midrule
        GMatching &0.007 &0.037 & 0.067& 0.028 & 0.062 & 0.156 & 0.233 & 0.113\\
        FSRL & 0.080 & 0.158 & 0.210 & 0.127 & 0.120 & 0.253 & 0.345 & 0.192\\
        MetaR  & .0.003& 0.235 & 0.293& 0.115 & 0.044 & 0.172 & 0.244 & 0.112\\
        \midrule
        FTAG (Random) & 0.228 & 0.416 & 0.525 & 0.331 & \textbf{0.191} & 0.479 & 0.641 & \textbf{0.325} \\
        
        FTAG (Time dependent) & \textbf{0.234} & \textbf{0.441} &\textbf{0.578} & \textbf{0.345} & 0.170 & \textbf{0.519} & \textbf{0.743} & 0.323 \\
        
        \bottomrule
    \end{tabular}
    
\caption{Hit@K results for one-shot learning on (i) one month of GDELT (Jan $2018$) and (ii) two years of ICEWS (Jan $2017$ - Jan $2019$) for relations in $\mathcal{T}_{meta-test}$. ``Random'' and ``Time dependent'' correspond to the query set selection method.}
\label{tab:icews_results}
\vspace{-0.2cm}
\end{table*}

\subsection{Loss Function and Training}
\label{sec:training}
\begin{algorithm}[t]
\caption{}
\label{alg:training}
\begin{algorithmic}

\Function{MakeTask}{$r, w, A_r, n_{shots}$} 
    \State $i \sim \textrm{Uniform}(1, |A_r|)$
    \State $S_r^t \gets \{A_r[i]\}$
    \State $limit \gets w + t_{j+1}$
    \While {$t_j < limit$}
        \State Add $A_r[j]$ to $Q_r^t$
        \State $j \gets j + 1$
    \EndWhile
    \State return $S_r^t$, $Q_r^t$
    
\EndFunction

\State \textbf{Input:}
\State 1: Meta training relations $\mathcal{T}$
\State 2: $\forall r, A_r = \{(s_i, r, o_i, t_i)\}$ \Comment{$t_i$ are sorted}
\State 3: Background TKG $G'$
\State 4: Episode length $w$, Number of shots $n_{shots}$

\For {$i=1,2,\ldots N$}
\State Shuffle relations in $\mathcal{T}$
\State Sample relation $r$ from $\mathcal{T}$
\State $S_r^t, Q_r^t \gets $ \Call{MakeTask}{$r, w, A_r, n_{shots}$} 
\State Sample $B^+$ from $Q_r^t$ and make $B^-$
\State $\mathcal{L} \gets \max (score^- - score^+ \lambda, 0)$
\State $\theta \gets \theta - \nabla \mathcal{L}$
\EndFor
\State return $\theta$
\end{algorithmic}
\end{algorithm}

For a given relation $r$ and its support set $S_r^t = \{(s_0, r, o_0, t)\}$, we have a set of positive quadruples ($Q_r^{t ^+}$) and construct the negative pairs ($Q_r^{t^-}$) by polluting the subject or object entities for each positive quadruple and the final query set is $Q_r^t = Q_r^{t^-} \bigcup Q_r^{t ^+}$. We want the positive quadruples to be close to the final representation of the support set and the negatives to be as far as possible. The objective function that we optimize is a hinge loss, defined as:
\begin{equation}
\mathcal{L} = max(score^- - score^+ + \lambda, 0 ) 
\end{equation}

The $score^+$ and $score^-$ are similarity scores calculated over $Q_r^{t ^+}$ and $Q_r^{t ^-}$. We employ episodic training over the task set $\mathcal{T}$ to optimize the loss function. Algorithm \ref{alg:training} summarizes the time dependent selection to construct the query set and the episodic training algorithm. 


\section{Experiments}
We evaluate our model on predicting new events for a relation by predicting the object entity $(s, r, ?, t)$ and conduct qualitative and quantitative experiments on   the model. This section includes the details of dataset and task construction, the proposed baselines, performance comparison and ablation studies on   variations of our model.

\subsection{Datasets}
{We use two datasets: Integrated Crisis Early Warning System (ICEWS) \cite{boschee2015integrated}  and Global Database of Events, Language, and Tone (GDELT) \cite{leetaru2013gdelt}. ICEWS and GDELT are two widely-used benchmarks for TKG completion tasks. Both are   large geopolitical event datasets automatically extracted and coded from news archives. In both datasets, the CAMEO-coding scheme is used to represent the events. CAMEO codes are a set of predefined geopolitical interactions that constitute knowledge graph relations. Each dyadic event is represented as a timed interaction (CAMEO code) between two geopolitical actors (knowledge graph entities). ICEWS is updated on a daily basis and GDELT every 15 minutes. From these datasets, we construct two new benchmarks for one-shot relational learning over TKGs.}

Our first dataset is constructed from two years of the ICEWS dataset, from Jan 2017 to Jan 2019. Events timestamps have daily granularity in ICEWS. We select the relations with frequency between 50 and 500 for the one-shot learning tasks and frequency higher than 500 as the background relations. Our second dataset includes one month of GDELT (Jan 2018). GDELT is much larger than ICEWS, since it is updated every 15 minutes, and the event timestamps have 15 minutes granularity. The low and high frequency thresholds for selecting tasks and background relations are 50 and 700, respectively, for the GDELT dataset. The rest of the dataset pre-processing is the same for both GDELT and ICEWS. Table \ref{tab:data} shows the statistics for both datasets.


\begin{table}[h!]
    \centering
    \begin{tabular}{c|c c c c}
    \toprule
        Dataset & $\#$ Ents & $\#$ Rels & $\#$ Tasks & $\#$ Quads\\
        \midrule
        ICEWS& 2419 & 153 & 66/5/14 & 7535\\
        GDELT& 1549 & 204 & 50/5/14 & 10420\\
    \bottomrule
    \end{tabular}
    \caption{Dataset statistics for two years of ICEWS($2017$-$2019$) and one month of GDELT (Jan 2018).$\#$Rels include all the meta relations and background relations, and $\#$Tasks is the number of relations in $\mathcal{T}_{meta-train}/\mathcal{T}_{meta-val}/\mathcal{T}_{meta-test}$.}
    \label{tab:data}
    \vspace{-1em}
\end{table}

\begin{table*}[t]
\centering
    \begin{tabular}{l || c | c | c || c c c |  c c c}
        \toprule
         & \multicolumn{3}{c|}{Setting} & \multicolumn{3}{c|}{GDELT} & \multicolumn{3}{c}{ICEWS} \\
        \cline{2-10}
        Model & Att & Rand & MatchNet & H@1 & H@10 & MRR & H@1& H@10 & MRR\\
        \midrule
        M1 &  &  &  & 0.045 & 0.225 & 0.114 & 0.060 & 0.558 & 0.197 \\
        M2 &  &  & \checkmark & 0.133 & 0.504 & 0.243 & 0.105 & 0.518 & 0.220\\
        M3 & \checkmark & \checkmark & \checkmark  & 0.197 & 0.535 & 0.293 & 0.123 & 0.616 & 0.245\\
        M4 & \checkmark &  & \checkmark & 0.169 & 0.491 & 0.265 & 0.138 & 0.654 & 0.269\\
        \midrule
        FTAG (Random) & \checkmark & \checkmark & &0.228 & 0.525 & 0.331 & \textbf{0.191} & 0.641 & \textbf{0.325}\\
        FTAG (Time dependent) & \checkmark & & &\textbf{0.234} & \textbf{0.578} & \textbf{0.345} & 0.170 & \textbf{0.743} & 0.323\\
        
        \bottomrule
    \end{tabular}
    
\caption{Ablation study on different components of the model (i) one month of GDELT (Jan 2018) and (ii) two years of ICEWS (Jan 2017 - Jan 2019) for relations in $\mathcal{T}_{meta-test}$.}
\label{tab:ablation}
\vspace{-0.3cm}
\end{table*}


\subsection{Baselines} There is no prior work on one-shot learning for temporal knowledge graphs. Therefore, we propose two different ways to evaluate our model:
\begin{enumerate}
    \item  \textbf{One-shot training of existing TKG models} To simulate the one-shot condition, we make a training set by adding all the quadruples of the background knowledge graph, as well as the quadruples of the \textit{meta-train}. Per each relation in the \textit{meta-test} and \textit{meta-val}, we also include exactly one quadruple into the training set. We test the model on the exact same quadruples from \textit{meta-test}. The TKG reasoning models used as baseline include: TADistMult \cite{garcia2018learning}, TTransE \cite{leblay2018deriving} and ReNet \cite{jin2019recurrent}.
    
    \item \textbf{FSL methods for static graphs:} We collapse the temporal training graph into an unweighted static graph. An edge exists between two entities in the static graph if there is a corresponding edge in the temporal graph at any time. We use three state of the art static low-shot learning methods: GMatching \cite{xiong2018one}, FSRL \cite{zhang2020few}, and MetaR \cite{chen2019meta}. Unlike the first two, MetaR doesn't incorporate any neighborhood information into its modeling, meaning that there is no difference between $(s, r, o, t_i)$ and $(s, r, o, t_j)$ in the test. In contrast, the one-hop neighborhood information provided for $(s, r, o, t_i)$ is different than $(s, r, o, t_j)$ during the test time of GMatching and FSRL.
\end{enumerate}{}


\subsection{Evaluation} We divide the dataset into train, validation, and test splits, as explained in Section \ref{sec:fewshot-setup} and visualized in Figure \ref{fig:split}. We evaluate the models using Hit@(1/5/10) and Mean Reciprocal Rank (MRR). We use all the entities in the dataset to generate a list of potential candidates for ranking. For each method, the model with the best MRR on the validation set is selected for evaluation over the test set. Table \ref{tab:icews_results} summarizes the results of prediction tasks on the ICEWS and GDELT datasets. The one-shot support example for $\mathcal{T}_{meta-test}$ is selected from the training period for the first set of baselines. For our method and the second set of baselines, it is possible to select a one-shot example from the test period. We also tested our method with the same one-shot example provided to the first three baselines, selected from the training period, but the difference was not significant. We run each method five times with a different random seed and report the average. Appendix \ref{sec:hyper} includes the details of hyperparameter selection and implementation.

{\noindent\textbf{Discussion}. We observe in Table \ref{tab:icews_results} that our model outperforms all the baselines. In particular, the improvement is significant for Hit@10. The episodic training used in our model provides more generalizability compared to the first set of baselines that use regular training. Our initial experiments show that the performance of these models are closer to the frequent relations and declines when evaluated over only sparse relations. Although the second set of baselines employ episodic training, these methods still fail to consider the temporal dependency between events, which is captured effectively by self-attention in our model.}

\subsection{Ablation Study}
{To demonstrate the importance of each component of our model, we conduct multiple ablation studies that evaluate the model from three main angles:
\begin{enumerate}
    \item The temporal neighborhood encoder added by self-attention to the model: We disable the sequential encoder and feed all the neighbors of an entity in $\{t-\ell, \dots  ,t-1 \}$ to the snapshot function $f_\eta$, as if they all happened at one timestamp (M1), shown in Table \ref{tab:ablation} as ``Att''.
    \item Query set selection method: According to Section \ref{sec:fewshot-setup}, it can either be random or time dependent. ``Rand'' is checked in Table \ref{tab:ablation} if the selection method is random, and time dependent otherwise. 
    \item We analyze the effectiveness of the inner product to a more complicated model and add Matching Network \cite{xiong2018one} to the query representation. It is shown as ``MatchNet'' in Table \ref{tab:ablation}.
\end{enumerate}
}

{Table \ref{tab:ablation} summarizes the ablation study's results, showing that the full pipeline of our proposed algorithm outperforms the other variations. M1 and M2 show the effectiveness of a sequential encoder, since disabling it cause a significant decline in the performance. It is worth noting that adding MatchNet helps to capture similarity information when the model is simple (M1 and M2). However, comparing FTAG/FTAG-R with M3 and M4 shows that, due to the lack of data, adding MatchNet will lead to overparameterizing the model and decreasing the performance, while self-attention is powerful enough to learn a representation that captures not only the temporal dependencies but also a similarity space that enables accurate prediction.}


\subsection{Performance Over Different Relations}
\label{sec:rel-analysis}
In this section, we conduct experiments to evaluate the model performance over each relation separately. Table~\ref{tab:relationAnalysis} shows relations in ICEWS test set by performance. Our model struggles on CAMEO codes ``1831'' and ``1823,'' which lie under a higher level CAMEO event \textit{``Assault''} coded as ``18.'' Also, we manually inspected the test examples for ``1823,'' for which ReNet performs very well. Our inspection shows that ReNet tends to generate higher ranks for a quadruple $(s, r, o, t)$ if it has already seen many examples of $s$, $o$ being paired  with any other relations. For example,
\textit{(ISIS, 1831, Afghanistan)} was the test example, and we found 30 matches for \textit{(ISIS, 183, Afghanistan)} in the training set. It is worth noting that ``1831'' is a subcategory of ``183'' in the CAMEO-code scheme. This was the case for 4 out of 5 query examples of ``1831.'' 
The one query example that ReNet doesn't perform well,  \textit{(ISIS, 1831, Libya)}, the combination of $ISIS$ and any other $Libya$ related entities only appeared 7 times in the training data. The rank predicted by our model for this query is 14, while the ReNet rank is over 1,000.
Our model doesn't use the information from the edges in the background graph. Although ReNet leverages this information, it could become biased toward them. Therefore, designing a few-shot model that leverages this information and is able to generalize well over new edges remains a challenge for future work. 

\subsection{Performance over Time}
\begin{figure}[ht]
    \centering
    \includegraphics[width=0.35\textwidth]{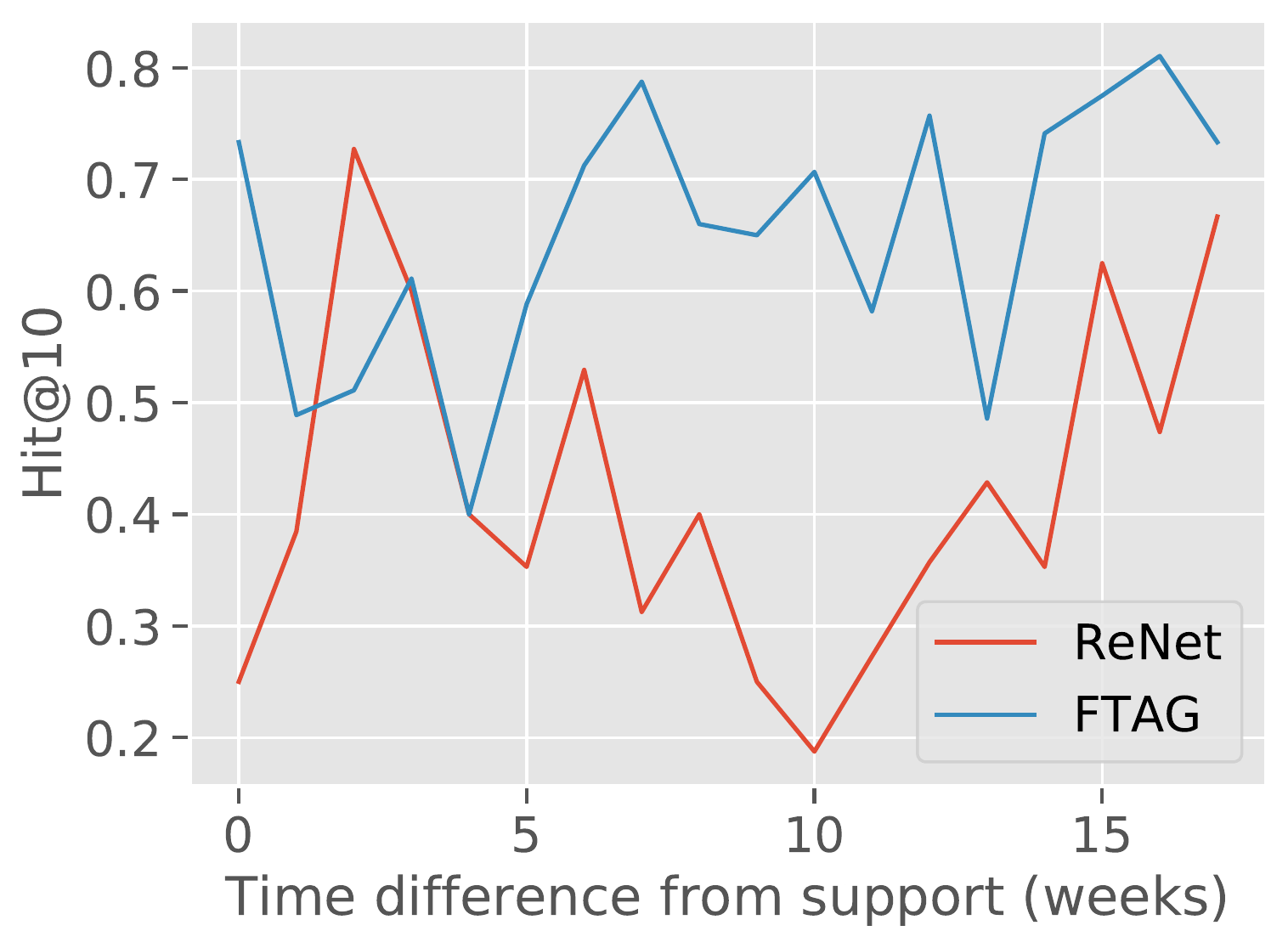}
    \vspace{-0.2cm}
    \caption{Model performance as the time between the prediction and the end date of the support set increases.}
    \label{fig:over_time}
    \vspace{-0.4cm}
\end{figure}

Figure~\ref{fig:over_time} visualizes the performance of our model over time for ICEWS dataset. Since relations selected for the task are very sparse, the number of query examples in one unit of time is very small. So we aggregated every 7 days. The y axis is the time difference between the query timestamp and its support example timestamp. Figure \ref{fig:over_time} shows that our model outperforms the best baseline over time. 

\begin{table}[t]
\centering
\vspace{0.3 cm}
    \begin{tabular}{l  c | c c }
        \toprule
         & &\multicolumn{2}{c}{Hit@10} \\
        \cline{3-4}
        CAMEO Code & Frequency &FTAG & ReNet \\
        \midrule

    1044 & 52 &0.300 & \textbf{0.600}\\
    1125 & 58 &\textbf{0.650} & 0.615\\
    1311 & 64 &\textbf{0.567} & 0.000\\
    186 & 71 &\textbf{0.533} & 0.250\\
    1831 & 97 &0.450 & \textbf{0.800}\\
    1122 & 121 &\textbf{0.757} & 0.667\\
    011 & 128 &\textbf{0.600} & 0.000\\
    0313 & 130 &\textbf{0.656} & 0.579\\
    1823 & 143 &0.133 & \textbf{0.286}\\
    1721 & 225 &\textbf{0.600} & 0.133\\
    0312 & 273 &\textbf{0.800} & 0.714\\
    063 & 283 &\textbf{0.688} & 0.364\\
    0333 & 292 &\textbf{0.612} & 0.353\\
    0332 & 348 &\textbf{0.785} & 0.463\\

        \bottomrule
    \end{tabular}
\caption{Hit@10 and MRR reported separately for each relation in the test tasks for ReNet and FTAG. The description of CAMEO codes is provided in Appendix~\ref{sec:cameo}.}
\label{tab:relationAnalysis}
\vspace{-0.5cm}
\end{table}

\section{Related Work}
Our work is particularly related to representation learning for temporal relational graphs, few-shot learning methods and recent developments of meta learning approaches for graphs. 

\noindent\textbf{Few-Shot Learning.} An effective approach for few-shot learning is based on learning a similarity metric and a ranking function using training triples~\cite{koch2015siamese,vinyals2016matching,snell2017prototypical,mishra2018simple}. Siamese networks~\cite{koch2015siamese} use a pairwise loss to learn a metric between input representations in an embedding space and then  use the learnt metric to perform nearest-neighbours separately.
Matching networks~\cite{vinyals2016matching} learn a function to embed input features  in a low-dimensional space and then use cosine similarity in a kernel for classification.  Prototypical  networks~\cite{snell2017prototypical} compute a prototype for each class in an embedding space and then classify an input using  the distance to the prototypes in the embedding. SNAIL~\cite{mishra2018simple}  uses temporal convolution to aggregate information from past experiences and causal attention layers to select important information from past experiences. Another paradigm of few-shot learning includes optimization-based approaches that usually include a neural network to control and optimize the parameters of the main network. One example is MAML \cite{finn2017model} that learns how to generalize with only a few examples and doing a few gradient updates. 

\noindent\textbf{Relation Learning for TKGs.} The temporal nature of TKGs introduces a new piece of information, as well as a new challenge in learning representation for TKGs. To model the time information \cite{leblay2018deriving, garcia2018learning, dasgupta2018hyte} embed the corresponding time by: encoding the time text by an RNN \cite{garcia2018learning}, low dimensional embedding vectors \cite{leblay2018deriving} and hyper planes \cite{dasgupta2018hyte}. Other methods try to capture the temporal entities' interactions by encoding them with a sequential model, such as \cite{trivedi2017know} that represents events as the point processes, and  \cite{jin2019recurrent} that aggregates the one-hop entity neighborhood at each timestamp by a pooling layer, and pass it to an RNN in an auto-regressive manner. 

\noindent\textbf{Few-shot Learning for Graphs.} Few-shot learning for graphs has recently gained attention. Xiong et al.~\cite{xiong2018one} pioneered and proposed a few-shot learning framework for link prediction over the infrequent relations. They extract a representation for each entity from its one-hop neighborhood and learn a common similarity metric space. A number   followup studies also combine local neighborhood structure~\cite{zhang2020few, du2019cognitive} or reasoning paths~\cite{wang2019meta, lv2019adapting} with a meta learning algorithm, such as MAML. Chen et al.~\cite{chen2019meta} leverage the same framework, although they don't use the local neighborhood structure. An adversarial procedure is used by Zhou et al.~\cite{zhang2020relation} to adopt features from high to low resource relations. Wang et al.~\cite{wang2019tackling} extend the effort from infrequent relations to tackle the unpopular entities problem in KGs by integrating textual entity descriptions.  Meta-Graph~\cite{bose2019meta} is a gradient-based meta learning approach for graphs. It leverages higher-order gradients along with a learned graph signature function to generates a graph neural network initialization. These approaches all assume a static graph. To the best of our knowledge, we are the first to study few-shot learning for temporal knowledge graphs. 


\section{Conclusion and Future Work}
We introduce a novel one-shot learning framework for temporal knowledge graphs to address the problem of infrequent relations in those graphs. Our model  employs a self-attention mechanism to sequentially encode temporal dependencies among the entities, as well as a similarity network for assessing the similarity between a query and an example. Our experiments demonstrate that the proposed method outperforms existing state-of-the-art baselines in predicting new events for infrequent relations. 
In future work, we would like to generalize the current one-shot learning to a few-shot scenario. 
Another direction is extending our framework to handle emergent entities, a challenge since new entities will have fewer interactions and thus considerably sparser neighborhood information.

\small
\bibliography{ref}
\clearpage
\appendix
\section{Attention Encoder Details}
\label{sec:app-att}
The Attention function used in Equation \ref{eq:attention} to calculate the attention score is called ``Scaled Dot Product Attention'' in \cite{vaswani2017attention} and defined as follows:
\begin{equation}
    \textrm{Attention}(Q, K, V) = softmax\big(\frac{QK^T}{\sqrt{d_k}}\big)V
\end{equation}

In order for the attention model to make use of sequential order, a positional encoding is added to the input embeddings. 
\begin{equation}
\begin{split}
    &PE_{(pos, 2i)} = sin(pos/10000^{2i/d_{model}}) \\
    &PE_{(pos, 2i+1)} = cos(pos/10000^{2i/d_{model}}),
\end{split}
\end{equation}

Where $pos$ is the position and $i$ is the dimension. The purpose of positional encoding is to introduce to the model the information about relative or absolute position of each element in the sequence. The positional encoding has similar dimension as $d_{model}$.

\section{Hyperparameters}
\label{sec:hyper}
\begin{figure}[h!]
    \centering
    \includegraphics[width=0.43\textwidth]{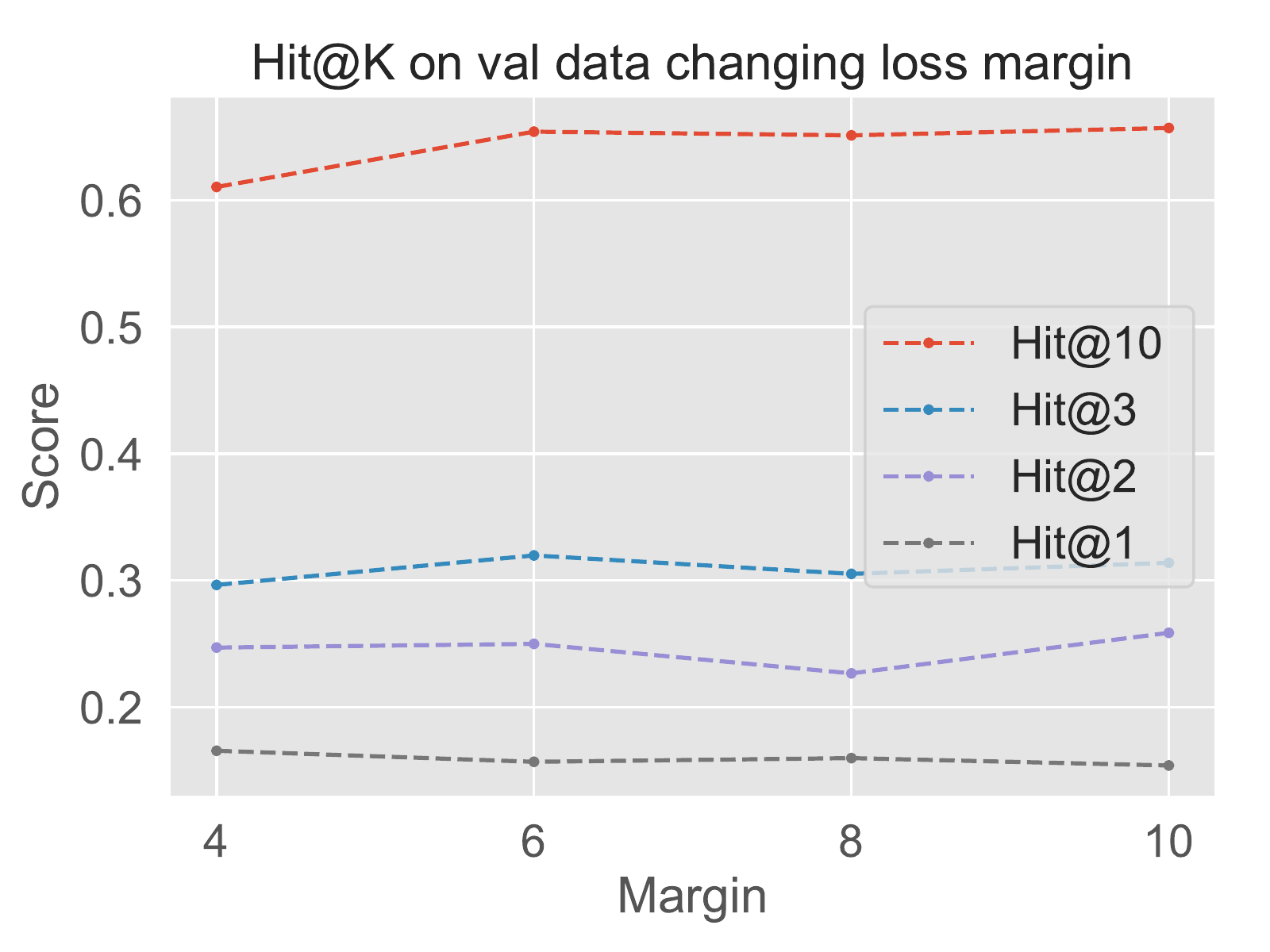}
    \caption{Model performance vs. the margin parameter.}
    \label{fig:margin}
\end{figure}

We select the relations with frequency between 50 and 500 for the one-shot learning tasks and frequency higher than 500 as the background relations for ICEWS dataset. The low and high frequency thresholds for selecting tasks and background relations are
50 and 700, respectively, for the GDELT dataset.The threshold for choosing the sparse relations should be selected such that the sparsity is preserved and also, we have enough data for training the model. We have selected the exact threshold values based on the prior work \cite{xiong2018one} which also is based on the above rationale. GDELT is less sparse than ICEWS, so we increased the upper threshold to increase the number of tasks for the training. \\
We use a manual tuning approach to select the model hyperparameters, during which we keep all the parameters constant except one and we run the model with the selected hyperparameters 5 times and select the best model over the validation set using MRR metric.\\
The episode length $w$ chosen to construct the datasets for one-shot learning from GDELT and ICEWS is 120 time units (e.g. 120 days for ICEWS). The history period is 20 days for ICEWS and 10 time units (every 15 minutes) for GDELT.
The embedding size for both datasets is $50$. We use one layer of multi-head attention with 4 heads. Number of heads is selected by hyperparameter search from 1 to 6. Attention inner dimension is 256. Attention parameters are similar for both datasets. The matching network performs 3 steps of matching. The loss margin is 10 for ICEWS and  18 for GDELT. We found out that increasing margin value affects the performance as it is depicted in Figure \ref{fig:margin}. The number of parameters for the model with this choice of hyperparameters is 1,380,656 for GDELT dataset and 1,469,056 for ICEWS. 
We use Adam optimizer with initial learning rate 0.001. 

\vspace{-2mm}
\section{Model Selection}
\label{sec:selection}
\begin{figure}[t]
    \centering
    \includegraphics[width=0.43\textwidth]{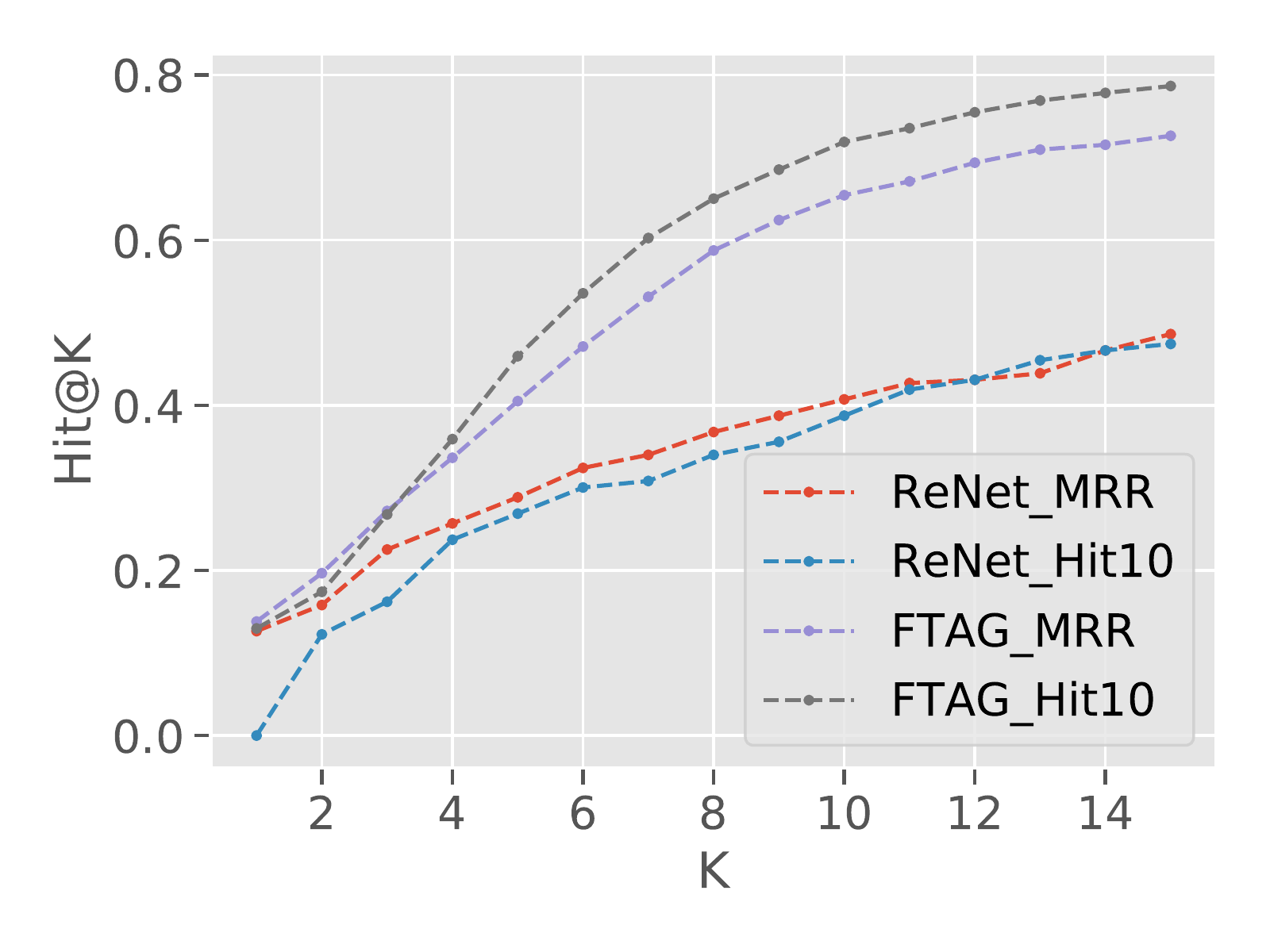}
    \caption{Hit@K for various values of K.}
    \label{fig:model-selection}
\end{figure}

During model selection, we noticed that there is a trade off between optimizing Hit@K for the smaller versus higher $K$. MRR favors smaller $K$ because smaller ranks contribute more to MRR. So using MRR for model selection results in a model with higher HIT@1. Figure \ref{fig:model-selection} show FTAG and ReNet performance for different $K$, with Hit@10 and MRR being used as the model selection metric. For ReNet, the validation set contains all the relations in the training set. This means that a model with the best Hit@10 doesn't necessarily give the best Hit@10 over $\mathcal{T}_{meta-val}$. We also tried using the same validation set as our model, that only contain $\mathcal{T}_{meta-val}$. This resulted in lower accuracy and higher variance for ReNet. 

\section{Model Analysis}
In this section we provide more insights on the shortcoming of the existing baselines and the justification about why our model outperforms these models. We compare our model against two categories of baselines: \\
\noindent\textbf{TKG baselines}: Regular TKG methods tend to get biased toward the frequent relations. We conducted some initial experiments to confirm it; We provided a training set containing all the relations to the model, and evaluated it on all the relations as well as sparse/frequent relations separately. The model performance (MRR/Hit@K) over all the relations was more close to the MRR/Hit@K for frequent relations and the MRR/Hit@K for sparse relations was much lower. The main difference between regular TKG models and our model is the episodic training framework, which enables our model to generalize well from only one example.
\begin{itemize}
    \item \textbf{TTransE/TATransE} are translation based models which are not able to handle one-to-many/many-to-one relations. they map the timestamp in a quadruple (s, r, o, t) into a lower dimensional space and are not capable of extrapolation (i.e. forecasting the future events).
    \item \textbf{ReNet}: Same as our model, ReNet generates a time-aware representation for an entity by aggregating the local neighborhood at each timestamp using a pooling layer and feeding it to an RNN. In Section \ref{sec:rel-analysis} we provide some insight on why and when the ReNet model outperforms our model. 
\end{itemize}
	
\noindent\textbf{FSL baselines}:The main difference between our model and static FSL models is a temporal neighborhood aggregator. Temporal adjacent events could convey useful information about the events that will happen in the future and different timestamps can have different effects on future events. The multi-head self-attention module in our model captures this information. We did some experiments on history length that indicated that as we increased the history length upto some point, it helped to improve the model performance. 

\begin{itemize}
    \item \textbf{GMatching} uses a mean pooling layer to aggregate the entities and edges adjacent to the given entity. 
    \item \textbf{FSRL} uses a weighted mean pooling layer with attention weights. The reason that FSRL works better than GMatching might be that a part of the temporal information is captured by attention weights.
    \item \textbf{MetaR} does not use the local neighborhood structure for extracting the embedding of a node. 
\end{itemize} 
To summarize, our model combines the benefits of  both approaches:  a self-attention to encode the temporal neighborhood information and  a temporal task definition for episodic training, resulting in better performance over the baselines.

\section{CAMEO Code Description}
\label{sec:cameo}
\begin{table}[!ht]
\centering
\vspace{0.3 cm}
    \begin{tabular}{ l l }
        \toprule
         Code &  Description \\
        \midrule

    1044 & Demand change in institutions, regime \\
    1125 & Accuse of espionage, treason  \\
    1311 & Threaten to reduce or stop aid \\
    185 & Assassinate \\
    1831 &Carry out suicide bombing \\
    1122 & Accuse of human rights abuses\\
    011 & Decline comment\\
    0313 & Express intent to cooperate on judicial matters)\\
    1823 & Kill by physical assault\\
    1721 & Impose restrictions on political freedoms\\
    0312 & Express intent to cooperate militarily  \\
    063 & Engage in judicial cooperation\\
    033 & Express intent to provide humanitarian aid  \\
    0332 & Express intent to provide military aid\\

        \bottomrule
    \end{tabular}
\caption{Mapping from CAMEO code to description.}
\label{tab:relation_description}
\end{table}

\section{Implementation Detail}
We implemented our solution using Pytorch. We run all the experiments on a CPU \texttt{Intel(R) Xeon(R) Gold 5220 CPU @ 2.20GHz}, and 53 GBs of memory. The \texttt{eval} function in \texttt{trainer.py} includes the details to calculate MRR and Hit@K metrics. The implementation and the dataset is available at \url{https://github.com/AnonymousForReview}

\section{Data Construction}
We provide the details of two newly constructed baselines for one-shot learning over temporal knowledge graphs. We conducted the following steps over both GDELT and ICEWS dataset: 
\begin{enumerate}
    \item A pre-processing step to deduplicate the dataset records by \texttt{Source Name} (subject), \texttt{Target Name} (object), \texttt{CAMEO Code} (relation), and  \texttt{Event Date} (timestamp).
    
    \item We divide the relations into two groups: frequent and sparse by their frequency of occurrence in the main dataset. Relations occurring between 50 and 500 in ICEWS, and 70 and 700 for GDELT are considered ``sparse.'' Those occurring more than 500 times in ICEWS and more than 700 times in GDELT are considered frequent.  
    
    \item The quadruples of the main dataset are then split into two groups based on their relations. The quadruples containing frequent relations make background knowledge graph kept in \texttt{pretrain.csv}, and the quadruples containing sparse relations are kept for meta learning process (meta quadruples) kept in \texttt{fewshot.txt}
    
    \item From the sparse relations, 5 are  selected for meta-validation, 15 for meta-test and rest kept for meta-training. 
    
    \item We split the meta quadruples into meta-train, meta-validation, and meta-test not only based on their relations, but also based on the non-overlapping time split explained in Figure \ref{fig:split} of the paper. 
\end{enumerate}

\subsection*{Data Format Description}
Each constructed dataset contains the following files:
\begin{outline}
    \1 \textbf{symbols2id.pkl}. A dictionary containing \texttt{ent2id}, \texttt{rel2id}, and \texttt{dt2id}, which are the mapping from entities, relations and dates to IDs respectively. 
    
    \1 \textbf{id2symbol.pkl}. A reverse mapping from IDs to symbols.
    
    \1 \textbf{data2id.csv}. A file containing all the quadruples after the deduplication step. The symbols are represented by their ids.
    
    \1 \textbf{pretrain.csv}. Contains the quadruples of the background knowledge graph. 
    
    \1 \textbf{fewshot.txt}. Contains the meta quadruples in text format. Each line is a tab separated quadruple with the order $s, r, o, t$.
    
    \1 \textbf{meta\textunderscore train.pkl}. A mapping from relations to meta quadruple IDs containing that relation. A quadruple ID indicates the line number corresponding to that quadruple in \textbf{fewshot.txt}. \textbf{meta\_test.pkl} and \textbf{meta\_val.pkl} are also created  similarly, using meta-validation and meta-test relations.
    
    \1 \textbf{hist\_l\_n}. A folder containing the entities' neighborhood information, with a maximum of $n$ neighbors at each snapshot and history length $l$. It includes the following files:
    
    \2 \textbf{hist\_o.pkl}. The object neighborhood of  meta quadruples in the \textbf{fewshot.txt}. The $i_{th}$ record corresponds to the quadruple in $i_{th}$ line of \textbf{fewshot.txt}.
    
    \2 \textbf{hist\_s.pkl}. The subject neighborhood of meta quadruples in the \textbf{fewshot.txt}. The $i_{th}$ record corresponds to the quadruple in $i_{th}$ line of \textbf{fewshot.txt}.
    
\end{outline}

.

\end{document}